\documentclass[10pt, conference, compsocconf]{IEEEtran}
\ifCLASSINFOpdf
\else
\fi
\IEEEoverridecommandlockouts 
\usepackage[numbers]{natbib}
\usepackage{multicol}
\usepackage[bookmarks=true]{hyperref}

\usepackage{graphics} % for pdf, bitmapped graphics files
\usepackage{epsfig} % for postscript graphics files
\usepackage{mathptmx} % assumes new font selection scheme installed
\usepackage{times} % assumes new font selection scheme installed
\usepackage{amsmath} % assumes amsmath package installed
\usepackage{amssymb}  % assumes amsmath package installed
\usepackage{algorithm}
\usepackage{algpseudocode}
\usepackage[table,xcdraw]{xcolor}

\usepackage[footnotesize, skip=0pt]{caption}
\usepackage{color}
\usepackage{svg}
\usepackage{siunitx}
\usepackage{url}
\usepackage{balance}

\usepackage[position=bottom,caption=false]{subfig}
\usepackage{dblfloatfix}
\usepackage{multirow}
\usepackage{graphicx}
\usepackage{animate}
\usepackage{tikz}
\usepackage{tikz-3dplot}
\usepackage{booktabs,tabularx}
\usepackage{bm}
\usepackage{comment}
\usepackage{mathtools}
\usepackage{placeins}
\usepackage[capitalise]{cleveref}
\usepackage{xr}
\usepackage{float}
\newcommand{\cblue}[0]{\color[HTML]{4A86E8}}
\newcommand{\cpink}[0]{\color[HTML]{FF00FF}}

\begin{document}
\title{Gradient-based Maximally Interfered Retrieval for Domain Incremental 3D Object Detection}

\author{\IEEEauthorblockN{Barza Nisar\IEEEauthorrefmark{1}, Hruday Vishal Kanna Anand\IEEEauthorrefmark{1}, Steven L. Waslander\IEEEauthorrefmark{2}}
\IEEEauthorblockA{Institute for Aerospace Studies \\
University of Toronto \\
Toronto, Canada \\}
\thanks{
% \dagger All authors are with University of Toronto Institute for Aerospace Studies,
% Toronto, Canada. 
\IEEEauthorrefmark{1} \{barza.nisar, vishal.kanna\}@mail.utoronto.ca}
\thanks{\IEEEauthorrefmark{2} steven.waslander@robotics.utias.utoronto.ca}
}


%Telephone: (800) 555--1212, Fax: (888) 555--1212}


\maketitle

\begin{abstract}
Accurate 3D object detection in all weather conditions remains a key challenge to enable the widespread deployment of autonomous vehicles, as most work to date has been performed on clear weather data.  In order to generalize to adverse weather conditions, supervised methods perform best if trained from scratch on all weather data instead of finetuning a model pretrained on clear weather data. Training from scratch on all data will eventually become computationally infeasible and expensive as datasets continue to grow and encompass the full extent of possible weather conditions. On the other hand, naive finetuning on data from a different weather domain can result in catastrophic forgetting of the previously learned domain. Inspired by the success of replay-based continual learning methods, we propose Gradient-based Maximally Interfered Retrieval (GMIR), a gradient based sampling strategy for replay. During finetuning, GMIR periodically retrieves samples from the previous domain dataset whose gradient vectors show maximal interference with the gradient vector of the current update. Our 3D object detection experiments on the SeeingThroughFog (STF) dataset \cite{dense} show that GMIR not only overcomes forgetting but also offers competitive performance compared to scratch training on all data with a \textbf{46.25\%} reduction in total training time.

\end{abstract}


% \fi
\IEEEpeerreviewmaketitle

\section{INTRODUCTION}

Human perception has the ability to learn incrementally and incorporate new information continuously while preserving previously learned knowledge. In contrast, standard artificial neural networks are known to suffer from catastrophic forgetting of previously seen data when learning on new data from a different domain. A common approach to alleviate forgetting is to retrain neural networks from scratch on all of the data each time a new domain is introduced. In many applications, the database will continue to grow and encompass new domains over time such that training on all of the data will eventually become prohibitive in terms of cost and computation time. In this perspective, continual learning (CL) enables a neural network to learn sequentially on data collected in new domains while retaining knowledge of previously learned domains \cite{de2021continual}. 

An important application that can benefit from continual learning is 3D object detection for autonomous driving in different weather conditions. Most large-scale autonomous driving datasets, such as Waymo \cite{waymo} and NuScenes \cite{nuscenes}, have been mainly collected in clear weather conditions. As a result, recent object detection models have mostly been trained on clear weather data. With a surplus of models trained only on clear weather data, the question yet to be answered is how these models can be fine-tuned on adverse weather data without forgetting their training on clear weather data. Recent developments in CL, as surveyed by De Lange et al. \cite{de2021continual}, show that it is possible to train a model sequentially on new data without forgetting old knowledge. Van de Ven et al. \cite{vandeven2019three} categorize CL scenarios into three types: 1) Task Incremental Learning (IL), 2) Domain IL, 3) Class IL. The task of learning on adverse weather domain without forgetting about previously learned clear weather data is a domain IL task. Although, domain IL approaches typically involve multiple sessions, where the model is finetuned on a new domain in each session, our experiments only show results for incremental learning from one domain to another. A related field in computer vision, called `Domain Adaptation (DA)', aims to utilize labelled data from a source domain to perform well on a different target domain for which labels are scarce. In this paper, we assume that sufficient labels are available for both source and target domains. Hence, our work is more closely related to domain IL. 
CL scenarios can be further split into offline and online settings. Online CL methods learn on a stream of samples (not-iid) observed only once as they arrive, whereas offline approaches process data in batches for many epochs. Offline learning is more common in the autonomous driving industry as it allows engineers more time to test and perfect the model before guaranteeing its safe deployment. Hence, our work considers the offline CL setting, where it assumes that the data from the new domain is available to train offline. Although offline learning allows more time to train, training from scratch on all LiDAR data can be very time consuming. For example, training a 3D object detector, such as PV-RCNN \cite{pvrcnn}, on full clear weather Waymo dataset using 4 T4 16Gb GPUs approximately takes around 8-9 days. On top of this, adding new domains to the dataset and tuning hyperparameters will significantly increase the total training time and consequently delay the final deployment of the model.

A prominent class of approaches used in CL, called \textit{rehearsal} or \textit{replay} based methods, train on a subset of samples from previous task/domain \footnote{"Task" and "domain" are used interchangeably.}, known as "replay samples", while learning new tasks to alleviate forgetting. Existing replay-based approaches are primarily designed for the online CL setting, whereby computations for replay sample retrieval \cite{MIR,GSS} or gradient projection \cite{AGEM} are executed every iteration. Such methods when directly employed in offline training offer little to no benefit over scratch training on all data as they consume relatively higher training time or memory while offering diminished performance.  Ideally, we want an efficient learning strategy that can improve performance on both old and new domains (compared to scratch training on individual domains), commonly known as positive backward and forward transfer, respectively. 

One of the primary goals of replay-based methods is to select replay samples for maximum reduction in forgetting \cite{MIR,GSS}. Inspired by Maximally Interfered Retrieval (MIR) \cite{MIR}, our work answers the question of which samples should be replayed to mitigate forgetting. While MIR is designed for the online CL setting, our work deals with offline learning.  The key idea behind MIR is to populate a mini-batch with previous task samples that suffer from an increase in loss given the estimated parameters update of the model.  Unlike MIR, our work, GMIR, periodically retrieves samples which exhibit maximal interference based on the \textit{gradient vector directions} compared to the gradient vector from the latest parameter update. While studies related to domain incremental learning for 2D object detection \cite{DIL2d} and class-incremental learning for indoor 3D object detection \cite{DACIL} exist, to the best of our knowledge, we are the \textit{first} to study offline domain incremental learning for efficient training of 3D object detectors in outdoor scenes. 
Our contributions can be summarized as follows:
\begin{itemize}
    \item We identify the task of offline domain incremental learning for efficient training of 3D object detectors on new domains. 
    \item We propose GMIR, a gradient-based sampling strategy for replay-based offline domain IL.
    \item Through our experiments on 3D object detection, we show that GMIR not only achieves zero forgetting but also improves performance on both old and new domains. Moreover, GMIR outperforms standard regularization and related replay methods while offering competitive performance with scratch training at \textbf{46.25\%} reduction in training time. Our code is available open-source\footnote{https://github.com/TRAILab/GMIR}.
\end{itemize}

\section{Related Work}

Continual learning approaches can be categorized into three major families depending on how task specific information is stored and used in learning new tasks. These include 1) regularization, 2) architectural or parameter isolation methods and 3) replay methods.

\subsection{Regularization}
Regularization methods introduce an extra regularization term in the loss function while training on the new data to penalize forgetting on old data. Zhizhong Li et al.~\cite{lwf} proposed knowledge distillation for class and task incremental learning, which only uses new task data to train the network while preserving the original capabilities.
%in which the previous model outputs, for new data inputs, are used as soft labels for previous tasks or classification layers. 
It has been shown that this strategy is vulnerable to domain shift between tasks \cite{expert_gates}. Another class of regularization methods estimates the importance of neural network parameters for old tasks. During training of later tasks, changes to important parameters are penalized. 
%(i.e. learning is slowed down for parts of the network important for previous tasks). 
Elastic Weight Consolidation (EWC) \cite{ewc} was the first work to establish this approach. Later, Synaptic Intelligence (SI) \cite{si} proposed a method to estimate importance weights online during task training.

\subsection{Architectural/Parameter Isolation}
Parameter isolation methods reserve different model parameters for each task to prevent any possible forgetting. Assuming no constraints on the architecture size, some methods introduce new branches for new tasks while freezing previous task parameters \cite{progressive}, or dedicate a model copy to each task \cite{expert_gates}. Another group of methods keep the architecture size constant but allocate fixed parts of the network, either parameters \cite{pathnet, packnet}, or units \cite{hardattention}, to each task. During training of the new task, previous task network components are masked out. Such methods, however, are limited by network capacity and require task ID to be known during prediction.

\subsection{Replay}
Replay-based methods store a subset of samples from previous tasks and replay these samples when learning a new task. These samples are either used as model inputs for joint training with new task samples \cite{icarl, er, tiny}, or to constrain the parameter update to stay in a feasible region which does not increase loss on previous samples \cite{GEM, AGEM}. 
Instead of joint training with previous task samples, GEM \cite{GEM} proposes to constrain new task updates by projecting its gradient direction onto the feasible region outlined by previous task gradients. A-GEM \cite{AGEM} reduces the update complexity by projecting the gradient on one direction estimated from a small collection of randomly selected replay samples. Both GEM and A-GEM allow for positive backward transfer. It was later shown by the authors of A-GEM that rehearsal on the buffer has competitive performance \cite{tiny}.

Recent works, which are more closely related to our work, address the crucial problem of how to populate the replay buffer. GSS \cite{GSS} aims to store diverse samples in the replay buffer by keeping samples whose gradient vectors exhibit small cosine similarity among each other. 
Instead of diversifying the memory buffer, MIR \cite{MIR} proposes to replay only those samples from memory which exhibit an increase in loss due to the estimated parameter update of the model. In terms of computation, for every iteration the method requires forward passing all previously stored samples twice per iteration to compute per sample losses before and after the estimated parameter update. Additionally, it requires two backward passes per iteration, one for the estimated parameter update and one for the final parameter update. Simplifying MIR for the offline setting, we propose replaying samples that show maximal gradient interference with the current gradient vector. While MIR chooses samples to replay in each minibatch, our gradient-based approach chooses maximal interfering samples to keep in the memory buffer for a fixed number of epochs before resampling. Hence, our approach is less computationally expensive and leads to a stable offline learning process.

\subsection{3D Object Detection}
3D object detection methods can be categorized by their sensor modality, with RGB cameras\cite{RGB}, LiDAR\cite{lidar} or a fusion of both\cite{fusion} being the most common. Although GMIR, by design, is not constrained by the input modality of the object detector, our experiments focus on LiDAR-based methods. LiDAR-based 3D object detectors are well established and remain state-of-the-art, with seminal works that include PV-RCNN\cite{pvrcnn}, VoxelRCNN\cite{voxel_rcnn}, PointRCNN \cite{pointrcnn}. These methods differ in the neural net architecture of the feature extraction backbone, the point cloud representations used (eg: voxel space, raw point clouds, abstraction such as pillars) and the number of detection stages, as surveyed by Zamanakos et al. \cite{lidar3dodsurvey}. In our work, we evaluate the effectiveness of GMIR on PointRCNN \cite{pointrcnn} and VoxelRCNN\cite{voxel_rcnn}, which employ different backbones and point cloud representations, to show the broad applicability of our technique.
\section{METHOD}
\begin{figure*}[ht]
\centering
\setlength{\abovecaptionskip}{-25pt plus 0pt minus 0pt}
\includegraphics[width=\linewidth] {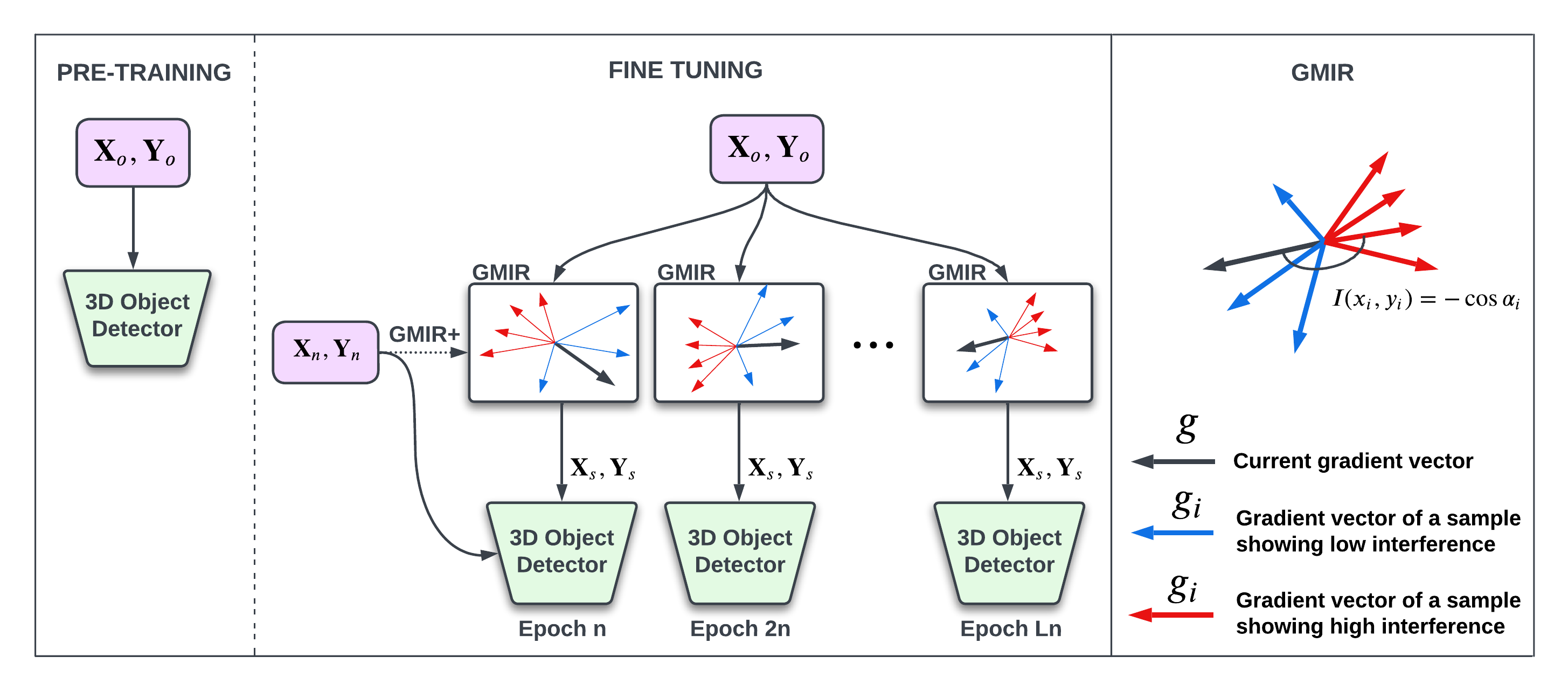}
\vspace{5pt}
\caption{In offline replay-based domain IL, a 3D object detector, pretrained on old domain dataset $\mathbf{D}_o$, is finetuned on new domain dataset $\mathbf{D}_n$ and a small subset of samples $\mathbf{D}_s$ chosen from the old dataset. During the finetuning stage, after every $n$ epochs, GMIR retrieves $\mathbf{D}_s$ by choosing top $K$ samples whose gradient vectors, denoted by red arrows, show maximal interference i.e. maximal $-\cos\alpha_i$ with the current gradient vector. It is to be noted that, for the sake of simplicity, gradient vectors are visualized as 2D vectors in this figure. In reality, gradient vectors have dimensionality equal to the number of parameters in the model.}
\label{fig:gmir}
\vspace{-10pt}
\end{figure*}
In the offline learning setting, our goal is to learn on data samples $\mathbf{D}_n = (\mathbf{X}_n, \mathbf{Y}_n)$ collected from a \textit{new} domain without interfering with previously learned samples $\mathbf{D}_o = (\mathbf{X}_o, \mathbf{Y}_o)$ representing another domain. One way to encourage learning without forgetting, is to learn a model, $f$, parameterized by $\theta$ that minimizes a predefined loss $\ell$ on new data as well as a subset of old data $\mathbf{D}_s = (\mathbf{X}_s, \mathbf{Y}_s) \subset \mathbf{D}_o$, where the size of $\mathbf{D}_s$ is fixed to $K$ samples. The key idea of our proposal is that we form $\mathbf{D}_s$ by finding samples in $\mathbf{D}_o$ whose gradient maximally interferes with the model's gradient vector from the latest update. For example, during the finetuning stage, if the gradient vector of the model points in the opposite direction of the gradient vectors computed at some of the old domain samples with respect to the current parameters, the model will update its parameters in a direction that will increase the loss on the old domain samples, resulting directly in forgetting. In order to mitigate forgetting on these samples, we include them in the training set. As the model's parameters change during finetuning, the subset of old domain samples which the model may be forgetting also change. Hence, we periodically resample the old domain samples to update the replay buffer with the latest interfering samples. Our method is described in Algorithm \ref{alg:1} and visualized in Figure \ref{fig:gmir}. More specifically, in the beginning of the finetuning stage, we first initialize $\mathbf{D}_s$ by randomly selecting $K$ samples from $\mathbf{D}_o$. After every $n$ epochs, we repopulate $\mathbf{D}_s$ with the top $K$ samples from $\mathbf{D}_o$ that exhibit the highest interference score. The interference score, $I$, for each sample $x_i, y_i$ in $\mathbf{D}_o$ after $t$ iterations is calculated by:

\begin{align}
    I(x_i, y_i) = - \frac{\left\langle g, g_i\right\rangle}{\|g\| \|g_i\|} \label{eq:I}\\
    g = \frac{\partial \ell\left(f\left(\mathbf{x}_t ; \theta_t\right), \mathbf{y}_t\right)}{\partial \theta} \\
    g_i = \frac{\partial \ell\left(f\left(x_i ; \theta_t\right), y_i\right)}{\partial \theta} \label{eq:gi}  
\end{align}
where $\mathbf{x}_t, \mathbf{y}_t$ is the minibatch used in the latest parameter update and $\langle \cdot{,}\cdot \rangle$ represents the dot product. In this formulation, the gradient, $g$, from the last iteration may be noisy. Alternatively, we can use the average gradient of all samples in $\mathbf{D}_n$ evaluated at the latest parameters $\theta_t$:
\begin{align}
    g = \frac{1}{N} \sum_{j=1}^{N} \frac{\partial \ell\left(f\left(\mathbf{x}_j ; \theta_t\right), \mathbf{y}_j\right)}{\partial \theta}
    \label{eq:gmir+}
\end{align}
where $N$ is the number of samples in $\mathbf{D}_n$. We call this variant GMIR+. It is to be noted that our method has three hyperparameters: 1) $D$: size of the old dataset $\mathbf{D}_o$ considered for sampling, 2) $K$: size of the replay buffer $\mathbf{D}_s$ i.e. number of samples selected for replay from $\mathbf{D}_o$, 3) $n$: number of epochs between each resampling of the replay buffer with GMIR.

\begin{algorithm}
\caption{GMIR Sample Selection}\label{alg:1}
\begin{algorithmic}[1]
\Require $g, f(\theta_t), \mathbf{D}_o$
\State $I_o = [\hspace{4pt}]$ \Comment{Interference score list for $\mathbf{D}_o$}
\For{$x_i, y_i$ in $\mathbf{D}_o$}
\State $\ell_i \gets \ell\left(f\left(x_i ; \theta_t\right), y_i\right)$ \Comment{Forward pass $x_i, y_i$}
\State $g_i \gets \partial \ell_i / \partial \theta$ \Comment{Equation~\ref{eq:gi}}
\State $I(x_i, y_i) \gets -  g, g_i / \|g\| \|g_i\|$ \Comment{Equation~\ref{eq:I}}
\State $I_o$.append($I(x_i, y_i)$)
\EndFor
\State $\mathbf{D}_s \gets$  SelectTopKSamples($I_o, \mathbf{D}_o$)

\end{algorithmic}
\end{algorithm}

 \section{Experiments}
\label{sec:exp}
% Barza
% Conducted experiments on Dense Dataset
\textbf{Dataset and Splits:} For 3D object detection experiments, we use the SeeingThroughFog (STF) dataset \cite{dense} which contains 12,994 LiDAR frames with 3D bounding boxes (Car, Pedestrian, Cyclist), collected under different weather conditions labeled as clear, snow, light fog and dense fog. We choose train, validation and test splits as 60\%, 15\% and 25\% respectively. Frames with fewer than 3000 points in the camera field of view are ignored. For training and testing, we create three splits: 1) \textit{clear}: contains `clear' weather frames, 2) \textit{adverse}: contains `snow', `dense fog' and `light fog' frames. 3) \textit{all}: combines frames from \textit{clear} and \textit{adverse} split. The \textit{adverse} training split $\mathbf{D}_n$ contains 3365 samples, whereas the \textit{clear} training split $\mathbf{D}_o$ contains $D=$ 3631 samples. 

\textbf{Evaluation setting}: For evaluation, we use the 3D object detection metrics defined in the KITTI evaluation framework \cite{kitti}. We report average precision (AP) at forty recall positions with a 3D IoU threshold of 0.7 for the most dominant class i.e. cars. 

\textbf{3D Object detection setup}: PointRCNN \cite{pointrcnn} and VoxelRCNN \cite{voxel_rcnn}, LiDAR-based 3D object detectors, are trained using OpenPCDet\footnote{https://github.com/open-mmlab/OpenPCDet} and their default training configuration. Training of each experiment is done on 4 T4 GPUs with a total batch size of 8 for 80 epochs. 

\textbf{Scratch Training Baselines}: In order to investigate the effectiveness of our approach, we compare our method to training from scratch on 1) \textit{clear}, 2) \textit{adverse}, and 3) \textit{all} weather splits.

\textbf{Finetuning Baselines}: In all finetuning experiments, we start training from the best model pretrained on \textit{clear} weather data and then finetune on the \textit{adverse} split. For replay-based methods, we jointly finetune on the \textit{adverse} split and a small subset of \textit{clear} split. We compare GMIR against several regularization and replay finetuning baselines described as follows: 1) Naive: Training is simply continued with the same hyperparameters as used in the pretrained model. 2) Lower Learning Rate (LR): Finetuning is done with a lower learning rate (0.003) compared to pretraining (0.01). The lower learning rate is chosen after tuning. 3) EWC \cite{ewc}: We set the weight ($\lambda$) for the regularization term to be 0.4, 4) MIR \cite{MIR}: The  top $K$ samples that exhibit highest increase in loss are retrieved \textit{every epoch} for replay. Retrieving samples every iteration, as in the original MIR can lead to infeasible training times in an offline learning setting. Every epoch, loss on a fixed buffer of $D$ clear weather samples is computed and compared with the loss on preceding epoch. Since the asymptotic overhead training time for modified MIR is $O$(num\_epochs $\times$ $D$), computing and storing losses for the entire clear weather training set (i.e. $D=$ 3631) every epoch will cost twice the training time of scratch training on the \textit{all} split. To achieve comparable training times with GMIR, we train MIR with $D=720$ samples. 5) A-GEM \cite{AGEM}: We implement two versions of replay buffer sampling: i) A-GEM: A-GEM with replay buffer of size $K$ populated with random samples from the \textit{clear} train split once before the start of training ii) A-GEM+: A-GEM with replay buffer of size $K$ randomly resampled from the \textit{clear} train split every $n=10$ epochs. 6) GSS \cite{GSS}: in order to populate the replay buffer with samples whose gradient vectors point in different directions, every $n$ epochs we compute the cosine similarity score between gradients of all clear weather samples ($D=3631$). We then rank the clear weather samples according to their maximum cosine similarity score and choose top $K$ samples with the smallest max cosine similarity scores. Computing and storing a very high dimensional gradient vector (order of millions) for each of the 3631 samples is infeasible due to limited memory. Hence, we approximate a gradient vector for each sample with respect to $1\%$ of randomly chosen model parameters. These selected parameters are kept same for all samples for a fair comparison. 7) Fixed Sampling: Simply finetuning on \textit{adverse} weather split and fixed $K$ samples randomly selected once from \textit{clear} weather training split. 8) Random Resampling: $K$ samples are randomly selected for replay every $n=10$ epochs. Both A-GEM+ and Random Resampling provide a fair comparison with GMIR since the resampling for all three is done at the same rate (i.e. every $n=10$ epochs). For all replay based methods the replay buffer size is kept the same i.e. $K=180$ (5\% of \textit{clear} weather train split of size $D$ = 3631 samples). \\
\textbf{Experiments for hyperparameter sensitivity analysis:} In order to analyse the effect of varying hyperparameters $(D, K, n)$ on the performance of GMIR, we repeat PointRCNN-GMIR experiments by varying one hyperparameter at a time while setting the other two to default values. The default values are $D$=100\% (i.e. 3631 samples), $K$=5\% (i.e. 180 samples), $n = 10$ epochs. For varying $D$, we randomly select and fix the predefined $D\%$ of samples from \textit{clear} train split at the beginning of each experiment and use these samples for retrieval of $\mathbf{X}_s, \mathbf{Y}_s$ throughout the training.

\begin{table}[]
\footnotesize
%\centering
\caption{Comparison of learning without forgetting methods for 3D object detection on STF\cite{dense}. We report 3D average precision (AP) of moderate cars on \textit{clear} and \textit{adverse} weather test split. mAP is the average performance on both splits. First 3 experiments are scratch training on \textit{clear}, \textit{adverse} and \textit{all} splits. The rest of the experiments are finetuning on \textit{adverse} split as explained in Section \ref{sec:exp}. {\color{blue}($\cdot$)}: Backward Transfer. {\cpink ($\cdot$)}: Forward Transfer. Higher values mean better performance.}
\label{tab:main}
\setlength\extrarowheight{-50pt}
\begin{tabular}{l|c|c|c}
    \toprule
     \textbf{Training}	&    \textbf{Clear AP} & \textbf{Adverse AP} & \textbf{mAP}  \\
     \textbf{method} &    \textbf{(Backward Transf.)} & \textbf{(Forward  Transf.)} &  \\
    \midrule 
    % \hline
    \multicolumn{4}{c}{\textbf{VoxelRCNN-Car}}    \\ \midrule
                                Clear & {\cblue 46.34} & 42.70 & 44.52 \\
                                Adverse  & 45.83 & {\cpink 45.47} & 45.65 \\
                                All & 46.79 & 46.41 & 46.60 \\ \hline
                                Naive & 45.88 {\cblue(-0.46)} & 46.54 {\cpink(+1.07)} & 46.21 \\
                                Low LR & 46.33 {\cblue(-0.01)} & 45.24 {\cpink(-0.23)} & 45.79 \\
                                EWC & 45.95 {\cblue(-0.39)} & 44.98 {\cpink(-0.49)} & 45.47 \\
                                MIR & 46.28 {\cblue(-0.06)} & 45.85 {\cpink(+0.38)} & 46.07 \\
                                A-GEM & 46.28 {\cblue(-0.06)} & 45.97 {\cpink(+0.5)}& 46.13 \\
                                A-GEM+ & 46.38 {\cblue(+0.04)} & 46.17 {\cpink(+0.7)} & 46.28 \\
                                GSS & 46.15 {\cblue(-0.19)} & 45.53 {\cpink(+0.06)}&  45.84 \\
                                Fixed Sampl. & 46.48 {\cblue(+0.14)} &  46.01 {\cpink(+0.54)} & 46.25  \\
                                Random Resampl. & 46.19 {\cblue(-0.15)} & 45.98 {\cpink(+0.51)} & 46.09  \\
                                GMIR (Ours) & 46.52 {\cblue(+0.18)} & \textbf{46.59 {\cpink(+1.12)}} & 46.56  \\
                                GMIR+ (Ours) & \textbf{47.34 {\cblue(+1.00)}} & 46.43 {\cpink(+0.96)} & \textbf{46.89} \\

    \midrule
    \multicolumn{4}{c}{\textbf{PointRCNN}} \\   \midrule
                                Clear & {\cblue 43.85}& 42.00 & 42.93 \\
                                Adverse  & 44.07 & {\cpink 43.12} & 43.60 \\
                                All & 45.74 & 44.23 & 44.99 \\ \hline
    Naive & 42.81 {\cblue (-1.04)} & 43.26 {\cpink(+0.14)} & 43.04 \\ 
    Low LR & 43.51 {\cblue (-0.34)} & 43.63 {\cpink(+0.51)} & 43.57 \\ 
    EWC & 43.06 {\cblue (-0.79)} & 43.06 {\cpink(-0.06)} & 43.06 \\   
    MIR & 43.85 {\cblue (0.00)} & 43.42 {\cpink(+0.30)} & 43.64 \\ 
    A-GEM & 43.94 {\cblue (+0.09)} & 43.45 {\cpink(+0.33)} & 43.70 \\ 
    A-GEM+ & 43.75 {\cblue (-0.10)} & 43.65 {\cpink(+0.53)} & 43.7 \\ 
    GSS & 43.21 {\cblue (-0.64)} & 43.39 {\cpink(+0.27)} & 43.23  \\ 
    Fixed Sampl. & 43.16 {\cblue (-0.69)} & 43.38 {\cpink(+0.26)} & 43.27 \\ 
    Random Resampl. & 42.72 {\cblue (-1.13)} & 43.19 {\cpink(+0.07)} & 42.96 \\ 
    GMIR (Ours) & \textbf{44.88 {\cblue (+1.03)}} & 43.81 {\cpink(+0.69)} & \textbf{44.35} \\ 
    GMIR+ (Ours) & 44.10 {\cblue (+0.25)} & \textbf{44.28 {\cpink(+1.16)}} & 44.19 \\ 
    \bottomrule
\end{tabular}
\end{table}
\begin{figure*}[ht]
\centering
\setlength{\abovecaptionskip}{-20pt plus 0pt minus 0pt}

\includegraphics[width=\linewidth]{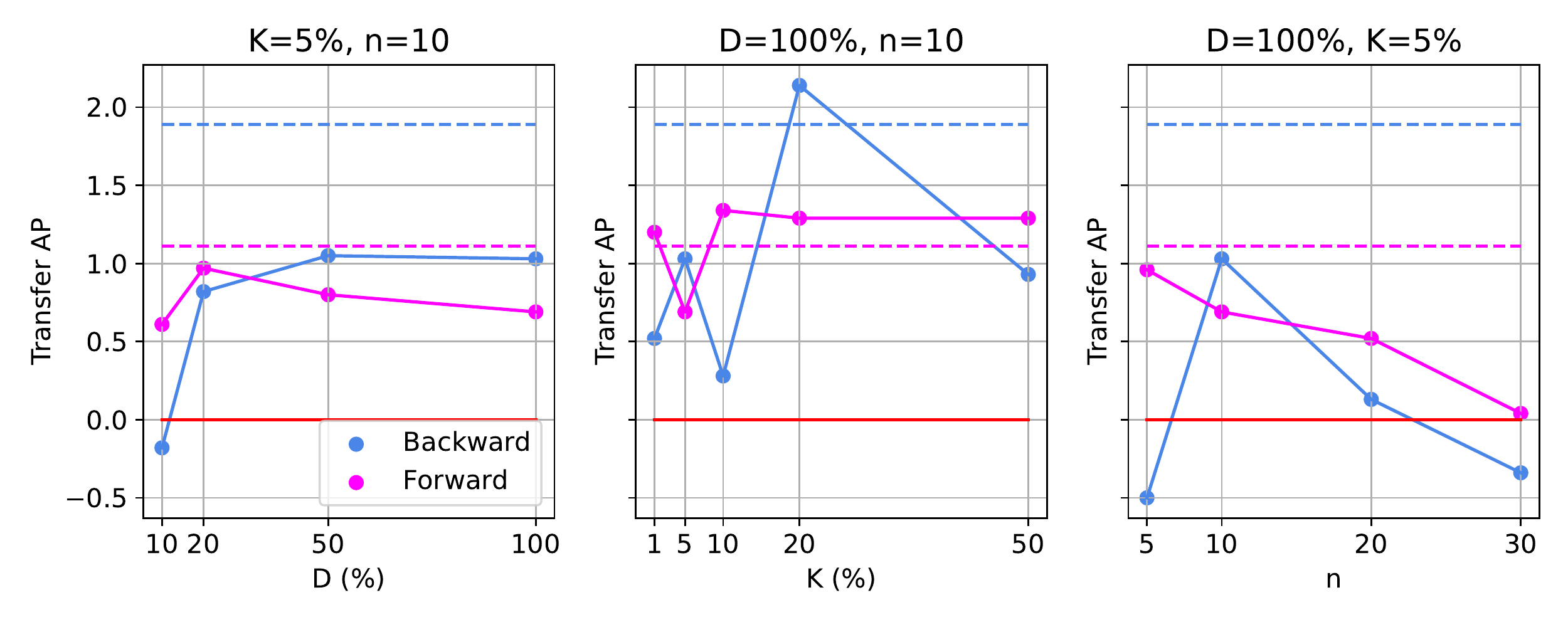}
\label{fig:D}

\caption{Effect of changing the hyperparameters of GMIR $(D, K, n)$ on PointRCNN performance.  {\cblue(---)}: Backward Transfer. {\cpink (---)}: Forward Transfer. Forgetting is indicated when backward transfer falls below zero (red line). {\cblue(- -)}: Performance of scratch training on \textit{all} data relative to the lower bound on \textit{clear} AP. {\cpink(- -)}: Performance of scratch training on \textit{all} data relative to the lower bound on \textit{adverse} AP.}

\label{fig:hyper}
\vspace{-10pt}
\end{figure*}

\begin{table}[]
\centering
\caption{Comparison of total training time of different replay methods applied to PointRCNN (in hours). Reduction in training time is reported with respect to total scratch training time on \textit{all}. $D$ represents the percentage of clear weather training set considered for sampling $K$ replay samples.}
\label{tab:2}
\begin{tabular}{|l|c|c|}
\hline
\textbf{Training Method}& \textbf{Training time (hrs)} &\textbf{ \% Reduction } \\ \hline
Scratch training on All & 16.0 & - \\
MIR($D=20\%$) & 13.4 & 16.25\\ 
A-GEM & 20.2 & - 26.25 \\
GSS & 11.2 & 30.0 \\
GMIR ($D=100\%$) & 11.6 & 27.5 \\ 
GMIR+ ($D=100\%$) & 14.9 & 6.88\\ 
GMIR ($D=20\%$) & \textbf{8.6} & \textbf{46.25} \\ \hline
\end{tabular}
\end{table}

\section{Results and Discussion}
\subsection{Backward and Forward Transfer on 3D Object Detection}
Table \ref{tab:main} shows 3D object detection results for the moderate car class on \textit{clear} and \textit{adverse} test splits. The AP results for scratch training on the individual \textit{clear} and \textit{adverse} splits serve as a lower bound on the performance we want to achieve with domain incremental learning, as they see none of the data from the other domain. The lower bound AP on \textit{clear} and \textit{adverse} are coloured as blue and pink, respectively. \textit{Backward} and \textit{Forward} transfer are calculated as the relative performance of finetuning experiments with respect to these lower bounds on old domain (\textit{clear}) and new domain (\textit{adverse}), respectively. From the results, we make the following observations:\\
\textbf{Naive finetuning:} As expected, naive finetuning on the \textit{adverse} split leads to forgetting on the \textit{clear} split and positive transfer on \textit{adverse} split.\\
\textbf{Regularization:} Simply lowering the learning rate (LR) reduces forgetting compared to naive finetuning but can sometimes slow down learning on the new domain as evident from the negative forward transfer in VoxelRCNN. Low LR, however, performs better than EWC on both domains. Although we did not tune EWC, we hypothesize that increasing EWC weight parameter for regularization term may reduce forgetting but it may also reduce forward transfer. \\
\textbf{Regularization vs Replay:} Baseline replay methods, i.e. MIR and A-GEM, result in positive forward transfer and reduced or zero forgetting compared to EWC. This is expected since replay methods have access to a subset of clear training samples.\\
\textbf{Replay baselines:} A-GEM, A-GEM+ and Fixed Sampling replay do sometimes exhibit positive backward transfer but their performance is not consistent between the two object detectors. Looking at the mAP of all baseline replay methods, A-GEM+ results in the highest mAP. \\
\textbf{Fixed Sampling vs Randomly Resampling:} Surprisingly, fixed buffer replay gives better performance than randomly resampling the buffer every 10 epochs. One would expect random resampling to increase the diversity of samples seen by the algorithm and hence lead to better generalization performance than fixed buffer replay. However, random resampling can potentially lead to a buffer which gives similar gradients to the adverse weather samples and hence may not be useful for learning. Also, learning on the same group of samples for more iterations leads to a more stable training which could explain higher performance of fixed buffer replay.\\
\textbf{GMIR/GMIR+ vs all baselines:} The results show that resampling the replay buffer periodically with maximally interfering samples throughout the training gives better performance than other baseline replay methods. Comparing GMIR to MIR, we see that resampling every 10 epochs instead of every epoch allows the model to iteratively learn on the same group of samples for more iterations, which results in better performance. Hence, GMIR is better suited to offline learning.
Since GMIR outperforms both A-GEM and A-GEM+, it shows that smart resampling of buffer is sufficient for both positive backward and forward transfer as opposed to constraining parameter updates. We also note that the performance of GSS could be limited by the approximation of gradient vectors from 1\% of model parameters. We leave the adaptation of GSS to offline learning setting to future work. Overall, both GMIR and GMIR+ not only overcome forgetting but achieve the highest positive backward and forward transfer compared to the baselines. Surprisingly, GMIR and GMIR+ also sometimes outperform scratch training on all data on one of the domains.

\subsection{Hyperparameter sensitivity analysis}
Figure \ref{fig:hyper} shows the effect of varying hyperparameters of GMIR (i.e. $D$, $K$, $n$) on the performance of PointRCNN. We report the performance of each experiment in terms of backward and forward transfer AP and make the following observations about each hyperparameter:\\
\textbf{Size of the old dataset $(D)$:} A striking revelation is that $D=20\%$ gives similar average performance compared to $D=50\%$ and $D=100\%$. Since LiDAR point clouds in autonomous driving datasets have redundancy in terms of similar looking objects in the road scenes, the plot shows that we don't necessarily need 100\% of the old dataset with GMIR to avoid forgetting. Hence, we are able to reduce the training time of GMIR without compromising on the performance by using only 20\% of the \textit{clear} train split. 
It is also note-worthy that increasing $D$ beyond 50\% does not increase the backward transfer, instead it decreases the forward transfer. Increasing $D$ increases the chances of GMIR retrieving different samples each time, which can be noisy for training and may lead to poor local minima for forward transfer. Therefore, at $D=100\%$, increasing the replay buffer size $K$, increases the chances of some samples being selected repeatedly and hence leads to more stable training. This could explain a high forward transfer at higher $K$ values for $D=100\%$. \\
\textbf{Size of the replay buffer $(K)$:} The best performing hyperparameter combination which results in the highest backward transfer for our setup is $K=20\%$, $D=100\%$ and $n=10$. With this combination, we can see that it is possible to surpass the performance of scratch training on \textit{all} denoted by the dotted lines. An interesting point to note is that even with 1\% of samples in the replay buffer, it is possible to achieve positive backward and forward transfer. \\ 
\textbf{Number of epochs $(n)$ between each resampling :} Performance of GMIR-based fine-tuning is sensitive to the number of epochs between each resampling. With $n$ set to a low number (in our case less than 10 epochs), GMIR does not help alleviate forgetting. Few epochs between each resampling results in few iterations to learn on each replay buffer before it gets resampled. This can lead to noisy updates and slow convergence to a better minima for backward transfer. If $n$ is set to a high number (more than 10 epochs in our case), we see a decline in both forward and backward transfer. This could be due to model overfitting to the replay buffer.

\subsection{Training time efficiency}
Table \ref{tab:2} presents a comparison of the total training time for different replay methods compared to the scratch training on \textit{all} data. Since our implementation of MIR requires forward passing $D=720$ samples (20\% of \textit{clear} train split) every epoch to keep track of per sample increase in loss on $\mathbf{D}_o$, MIR with $D=20\%$ takes longer to train then GMIR with $D=100\%$. The lower training time of GMIR is due to the less frequent update of replay buffer compared to MIR.
A-GEM takes the longest to train. This is because every iteration A-GEM requires loading a \textit{clear} batch separately in addition to the \textit{adverse} batch which can be time consuming. Moreover, training time for A-GEM increases with the number of times gradient projection is required. Worst-case time complexity for A-GEM is in the order of total number of training iterations, assuming gradient projection is required in every iteration. For GMIR, the training time increases with $D$ i.e. the size of the old dataset $\mathbf{D}_o$ considered in sample retrieval and the number of times resampling is done. GMIR+ takes longer to train compared to GMIR since in every resampling stage, it has to additionally loop over the new dataset to compute average gradient (see eq. \ref{eq:gmir+}). While GMIR on $D$=100\% of clear training set offers 27.5\% reduction in training time compared to scratch training, GMIR on $D$=20\% results in 46.45\% reduction in training time while still offering positive backward and forward transfer and similar mAP over both domains compared to $D$=100\%. 
Comparing the training times with the average precision results for all methods, we can conclude that GMIR provides the best performance and time tradeoff. 

\section{Limitations}
Although GMIR reduces the training time compared to scratch training on all of the data, our implementation of GMIR calculates the interference score one sample at a time. A more efficient implementation of GMIR can be to parallelize interference score calculation across multiple GPUs which will significantly reduce the overhead time. The main limitation of our method is that it assumes we have a subset of old domain dataset in memory to retrieve samples from. The size of the old dataset will grow as we include data from new domains. Hence, storing old dataset in memory can be infeasible due to storage constraints. Although, we have shown that GMIR works well even with 20\% of old data in memory, we can further try to bound the old dataset size by storing only those samples that were replayed most frequently in the previous training. Moreover, GMIR is sensitive to the tuning of hyperparameters $K$ and $n$, which may be dataset and model dependent.

\section{CONCLUSION}
We have proposed a strategy to retrieve important samples from previous domains which can be used for fine-tuning the previous model on new domains for domain incremental 3D object detection. Through 3D object detection experiments, we have shown that fine-tuning with maximal interfering previous domain samples overcomes forgetting and improves performance on both old and new domains, outperforming random sampling and standard baselines. Our method offers an efficient alternative to scratch training on all data.

\bibliographystyle{IEEEtran}
\bibliography{references}

\end{document}